\documentclass[12pt]{iopart}

\expandafter\let\csname equation*\endcsname=\relax
\expandafter\let\csname endequation*\endcsname=\relax

\usepackage{amsmath}
\usepackage{amssymb}
\usepackage{amsthm}
\usepackage{accents}

\usepackage{subfigure}
\usepackage{graphicx}
\usepackage{xcolor}

\usepackage{tabularx}
\usepackage{booktabs}
\usepackage{multirow}

\begin{document}

\title[]{Machine Learning Approach to Force Reconstruction in Photoelastic  Materials}

\author{Renat Sergazinov}
\address{Department of Statistics, Texas A\&M University, Blocker Building, 155 Ireland St, College Station, TX 77843, USA}
\ead{mrsergazinov@tamu.edu}

\author{Miroslav Kram\'ar}
\address{Department of Mathematics, University of Oklahoma, 601 Elm Avenue, Norman, OK 73019, USA}
\ead{miro@ou.edu}

\begin{abstract}
Photoelastic techniques have a long tradition in both qualitative and quantitative analysis of the stresses in granular materials. Over the last two decades, computational methods for reconstructing forces between particles from their photoelastic response have been developed by many different experimental teams. Unfortunately, all of these methods are computationally expensive. This limits their use for processing extensive data sets that capture the time evolution of granular ensembles consisting of a large number of particles. In this paper, we present a novel approach to this problem that leverages the power of convolutional neural networks to recognize complex spatial patterns. The main drawback of using neural networks is that training them usually requires a large labeled data set which is hard to obtain experimentally.  We show that this problem can be successfully circumvented by pretraining the networks on a large synthetic data set and then fine-tuning them on much smaller experimental data sets.  Due to our current lack of experimental data, we demonstrate the potential of our method by changing the size of the considered particles which alters the exhibited photoelastic patterns more than typical experimental errors.
\end{abstract}

\section{Introduction}
\label{sec::intro}
Granular materials consist of macroscopic particles, they are ubiquitous in nature and indispensable for a large variety of human activities. Powders and grains play a crucial role in many applications such as agriculture, construction, and chemical industry. As noted in \cite{duran}, the worldwide annual production of grains and aggregates of various kinds exceeds ten billion metric tons. Moreover,  granular materials are the second most manipulated materials in the industry after water. So, understanding the physics of granular materials has a major economic impact. 

\begin{figure}
    \centering
    {\includegraphics[width = \textwidth]{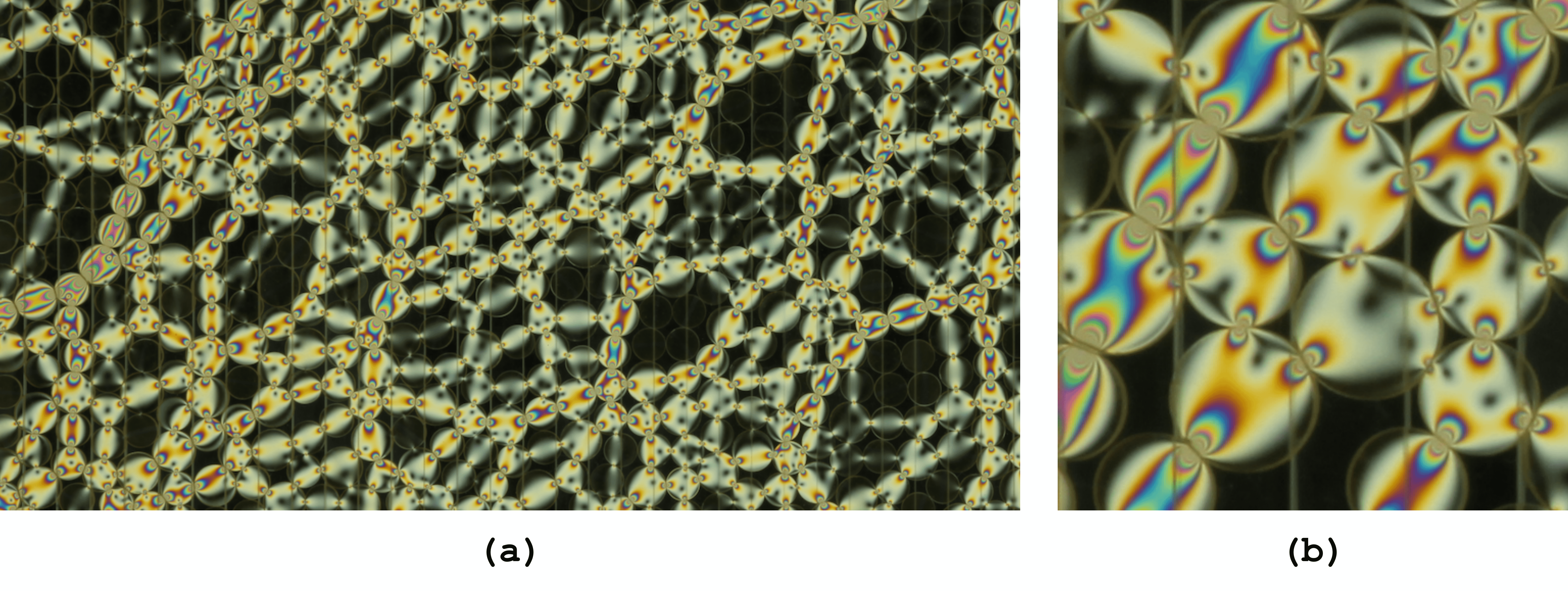}}
    \caption{[Courtesy of D. Wang.] 
    (a) An example of a photoelastic response of particles in a sheared system~\cite{wang2018microscopic} shows that the stresses (forces) in the material propagate along bright filamentary structures called force chains.  (b) Detail of the photoelastic pattern on individual particles. Similar patterns are produced by using monochromatic light and they can be used to reconstruct forces acting between the particles.} 
    \label{fig::exp}
\end{figure}

Despite extensive efforts by scientists, properties of granular systems are still not well understood and some of them remain rather obscure. However, it is well known that the forces in these systems do not propagate uniformly~\cite{bouchaud1998models, majmudar05a, clark2015nonlinear} but along highly anisotropic filamentary structures called force chains, shown in Fig.~\ref{fig::exp}.  The structure of the force chains is crucial for revealing the underlying physical causes of a wide variety of phenomena \cite{falcon2004nonlinear, dorbolo2002electrical, jia_prl99,makse_pre04, lherminier2014revealing} and plays an important role in determining the mechanical properties of granular systems.  The first attempts to experimentally study stresses in granular systems by using photoelastic materials  were conducted by Wakabayashi~\cite{wakabayashi1950photo} and Dantu~\cite{dantu1957proceedings}.   Later, Behringer and Majmudar used photoelastic methods~\cite{majmudar05a} to estimate the forces acting between the particles. We refer the reader to~\cite{daniels2017photoelastic} for a comprehensive review of photoelastic methods and their applications to granular materials. 

The visualization of the force chains, depicted in  Fig.~\ref{fig::exp}(a), is produced by placing an ensemble of photoelastic particles between two polarizing filters. The fact that the degree of birefringence at each location in the photoelastic material depends on local stress produces a visual pattern of alternating bright and dark fringes, visible in  Fig.~\ref{fig::exp}(b). The precise pattern depends in a complex manner on the orientation of the polarizers, the shape of the material, and how it is stressed. If a circular particle is illuminated by monochromatic light, then its photoelastic response can be computed from the forces acting on it, see Section~\ref{sec::photoelasticity} for more details. However, to determine the forces from the pattern one needs to solve a much more complicated inverse problem. 

Several computational methods for reconstructing the forces between circular particles were implemented by different groups~\cite{codePEDiscSolve,codeGrainSolver}. While the implementations differ, the algorithms are based on the general ideas presented in~\cite{majmudar05a}  and they roughly proceed as follows. First, the positions of the particles are calculated and the force-bearing contacts between them are identified. Then, for each particle the forces acting on it at the force-bearing contacts are guessed. Finally, an optimization algorithm is used to minimize the difference between the experimentally observed pattern exhibited by the particle and the pattern generated from the reconstructed forces.

As mentioned in~\cite{daniels2017photoelastic}, the positions of the particles can be identified with high accuracy using the Hough transform. However, there might be large discrepancies when it comes to force reconstruction. The reconstructed force between two particles can considerably vary depending on which of the two particles is used for the reconstruction.  Moreover, a computationally expensive evaluation of the photoelastic response has to be performed in every iteration of the optimization process. In this paper, we use machine learning to accurately reconstruct the forces acting on a particle from its photoelastic response.

Machine learning algorithms are very efficient for finding and recognizing patterns in complex data~\cite{bishop2006pattern, lecun2015deep}. They are extensively used in speech and image recognition as well as predictive analysis,  search systems, data visualization, and anomaly detection~\cite{lecun2015deep, besacier2014automatic, chorowski2016towards, jiao2019survey}.  In particular, convolutional neural networks (CNN) are well suited for the recognition of spatial patterns~\cite{lecun2015deep, krizhevsky2012imagenet} such as those exhibited by the photoelastic particles. 

\begin{figure}
    \centering
    {\includegraphics[width = \textwidth]{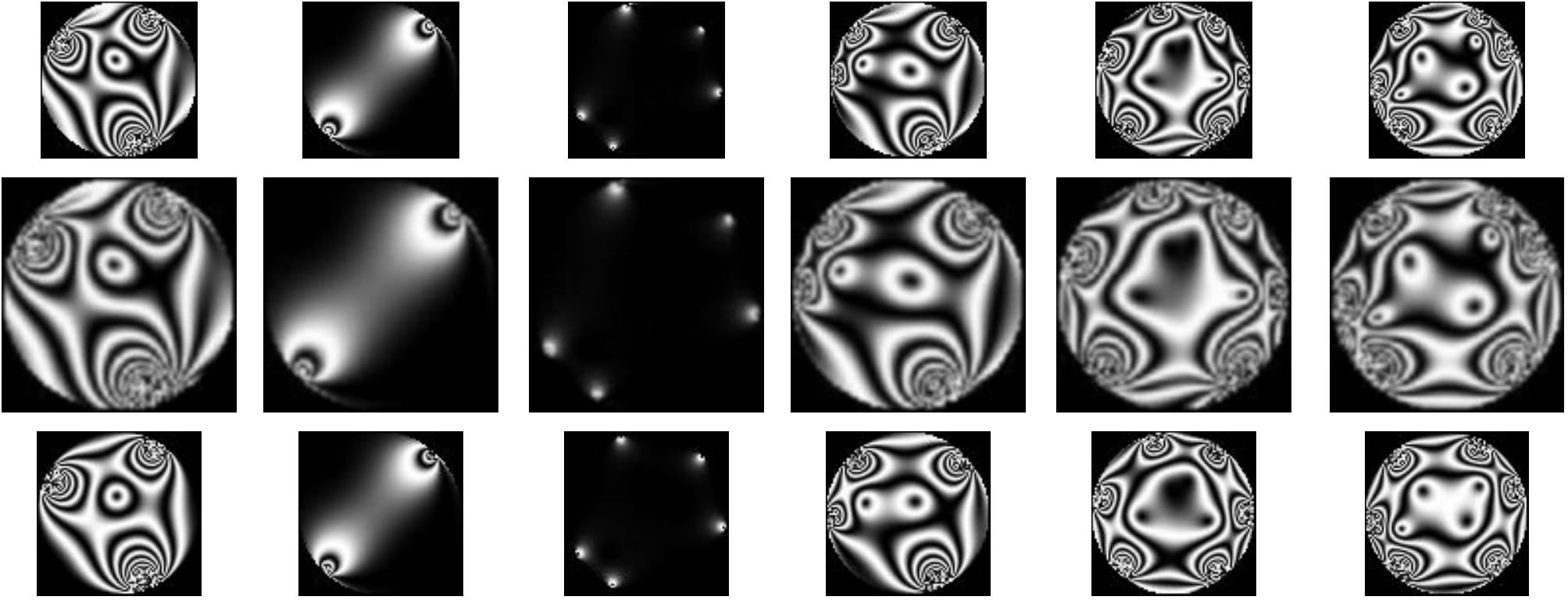}}
    \caption{Top row: Computer-generated photoelastic responses on a single particle for a variety of different forces. Middle row: The patterns produced from the images in the top row by resizing them to $128\times 128$ pixels using the nearest neighbor interpolation and then smoothing them by applying Gaussian blur with $\sigma = 1$. Bottom row: Computer-generated photoelastic response based on the forces reconstructed by a CNN from the preprocessed particle images depicted in the middle row.}
    \label{fig::true_vs_pred}
\end{figure}

We show that CNNs produce accurate force reconstruction on synthetic data. This is visually demonstrated by Fig.~\ref{fig::true_vs_pred}. The patterns depicted in the top row are computer-generated photoelastic responses of a single particle subject to a variety of different (known) forces acting on it.  The images in the middle row are created by first resizing the particles to bring out smaller details and then applying a Gaussian filter to make them look more natural, see Section~\ref{sec::training} for more details. Note that the corresponding images in the top and bottom row are quite similar.  While the precise results are presented in Section~\ref{sec::results}, we just mention that the mean absolute error in predicted positions of the force impact points is around two pixels and the mean absolute percentage error of the predicted force magnitudes is approximately  $10\%$.

To achieve high accuracy of the force reconstruction we experimented with diverse architectures of CNNs.  In particular, we considered three different models: VGG19,  InceptionResNetV2,  and  Xception,  each pre-trained  on  the  ImageNet  data  set~\cite{simonyan2015a, szegedy2017inception, chollet2017xception}.  Despite the high accuracy of our models trained on synthetic data, we cannot guarantee that their predictions on the experimental data would be equally accurate.  Unfortunately, training the models directly on experimental data is rather unrealistic because the required number of labeled samples is too large to produce experimentally.  To overcome this problem we suggest to pre-train the models on synthetic data and then transfer them to the experimental data by using a relatively small data set. Such a method has already been successfully used for different tasks such as object detection, optical flow estimation,  and scene understanding \cite{tremblay2018training, handa2015scenenet, mayer2016large}. Currently, we do not have the necessary experimental data to prove our claim. Thus, we demonstrate the potential of this approach by transferring our pre-trained model to smaller-sized particles, which we believe should serve as a proxy for the worst-case difference between experimental and synthetic data.  

The paper is organized as follows. Section~\ref{sec::photoelasticity} contains a brief overview of photoelastic theory and explains how the photoelastic response on a circular particle is computed from the forces acting on it.  In Section~\ref{sec::reconstruction} we formalize the problem of force reconstruction and summarize earlier works concerned with solving this problem. The machine learning framework for the force reconstruction is discussed in Section~\ref{sec::MLReconstruction} and the CNN models defined there are trained using the protocol outlined in Section~\ref{sec::training}.  Finally, our results are presented in Section~\ref{sec::results}. We conclude the paper with some final remarks in Section~\ref{sec::conclusions}.

\section{Photoelastic Theory} 
\label{sec::photoelasticity}

Photoelastic methods are widely used for quantitative analysis of stress in granular media. In this section, we provide a brief overview of photoelasticity while a detailed treatment can be found in \cite{frocht1946photoelasticity}. A comprehensive review of the experimental methods for granular media based on photoelastic theory is presented in  \cite{daniels2017photoelastic}. Figure~\ref{fig::true_vs_pred} shows a pattern of dark and bright fringes formed by illuminating photoelastic particles placed in between two polarizing filters. The intensity field $I(x,y)$ of the pattern depends on local stresses inside the particle. To be more precise, let us consider the stress tensor inside the particle 
$$
\sigma(x,y) = \begin{pmatrix}
\sigma_{xx}(x,y) & \sigma_{xy}(x,y) \\
\sigma_{xy}(x,y) & \sigma_{yy}(x,y) 
\end{pmatrix}.
$$
We denote its eigenvalues (principal stresses) by 
$$
\sigma_{\pm}(x,y) = \frac{1}{2}\left( \sigma_{xx}(x,y) + \sigma_{yy}(x,y) \pm \sqrt{(\sigma_{xx}(x,y) + \sigma_{yy}(x,y))^2 - 4 \sigma_{xy}^2(x,y)}\right).
$$
The intensity field exhibited by a particle is then given by 
\begin{equation}
\label{eqn::photpresponse}
I(x,y) =I_0\sin^2 \frac{\pi(\sigma_{+}(x,y)  - \sigma_{-}(x,y)) h C(\lambda)  }{\lambda},    
\end{equation}
where $h$ is the material thickness, $\lambda$ is the wavelength of light, $C$ is the stress optic coefficient dependent on $\lambda$, and $I_0$ is a normalization constant related  to the intensity of the light. In this paper, we consider $8$-bit grayscale digital images, $I_0 = 255$ and  $h C(\lambda)/\lambda = 0.18$ to closely match the experimental images used for calibration in \cite{majmudar2006experimental}.

The stress tensor can be computed from the forces acting on the particle using elasticity theory~\cite{landau1986lifshitz,love2013treatise}. 
We briefly explain how the tensor is computed on a single particle while a detailed treatment in the case of three forces can be found in~\cite{chau2018analyzing}.  Due to the nature of the problem, we can suppose that the Saint-Venant's compatibility condition is satisfied, i.e.,  there are no gaps or overlaps in the material. In the absence of internal body forces, this condition can be formulated by the following equation
\begin{equation}
\label{equn::compatibility}
    \frac{\partial^2\epsilon_{xx}}{\partial y^2} -2 \frac{\partial^2\epsilon_{xy}}{\partial x\partial y} + \frac{\partial^2\epsilon_{yy}}{\partial x^2} = 0,
\end{equation}
where $\epsilon$ is the strain tensor. To write the above equation in terms of stress we use the generalized Hooke's law that imposes the following linear strain stress relation:
\begin{equation}
\label{eqn::hook}
    \epsilon_{xx} = \frac{\sigma_{xx} - \nu \sigma_{yy}}{E}, \quad \epsilon_{yy} = \frac{\sigma_{yy} - \nu \sigma_{xx}}{E},\quad  \epsilon_{xy} = \frac{(1+\nu)\sigma_{xy}}{E}, 
\end{equation}
where $\nu$ is the Poisson's ratio and $E$ is the Youngs's modulus. The fact that the particle is in mechanical equilibrium implies 
\begin{equation} 
\label{eqn::equilibrium}
\begin{split}
\frac{\partial \sigma_{xx}}{\partial x} + \frac{\partial \sigma_{xy}}{\partial y} & = 0, \\
 \frac{\partial \sigma_{yy}}{\partial y} + \frac{\partial \sigma_{xy}}{\partial x} & = 0.
\end{split}
\end{equation}
To make sure that the stress tensor satisfies the equilibrium condition~(\ref{eqn::equilibrium}) we search for a solution in the form of the Airy stress function $\varphi$ and express the components of the tensor $\sigma$ as 
\begin{equation}
    \label{eqn::Airy}
    \sigma_{xx} = \frac{\partial^2 \varphi}{\partial y^2}, \quad \sigma_{yy} = \frac{\partial^2 \varphi}{\partial x^2}, \quad \quad \sigma_{xy} = \frac{\partial^2 \varphi}{\partial x \partial y}.
\end{equation}
By combining  Equations~(\ref{eqn::hook}) and (\ref{eqn::Airy}) we get an expression for the strain tensor in terms of $\varphi$. We substitute this expression into Equation~(\ref{equn::compatibility}) and obtain the final equation 
\begin{equation}
    \label{eqn::FinalElastic}
    \left( \frac{\partial^2}{\partial x^2} + \frac{\partial^2}{\partial y^2} \right)\left( \frac{\partial^2\varphi }{\partial x^2} + \frac{\partial^2 \varphi}{\partial y^2} \right) =0 .
\end{equation}
As is typical for partial differential equations we need to impose the correct  boundary conditions. In our case, it is natural to require that there is no stress on the boundary of the particle except at the points where the external forces act on it. 

\begin{figure}
\label{fig::sterss}
    \centering
    {\includegraphics[width = \textwidth]{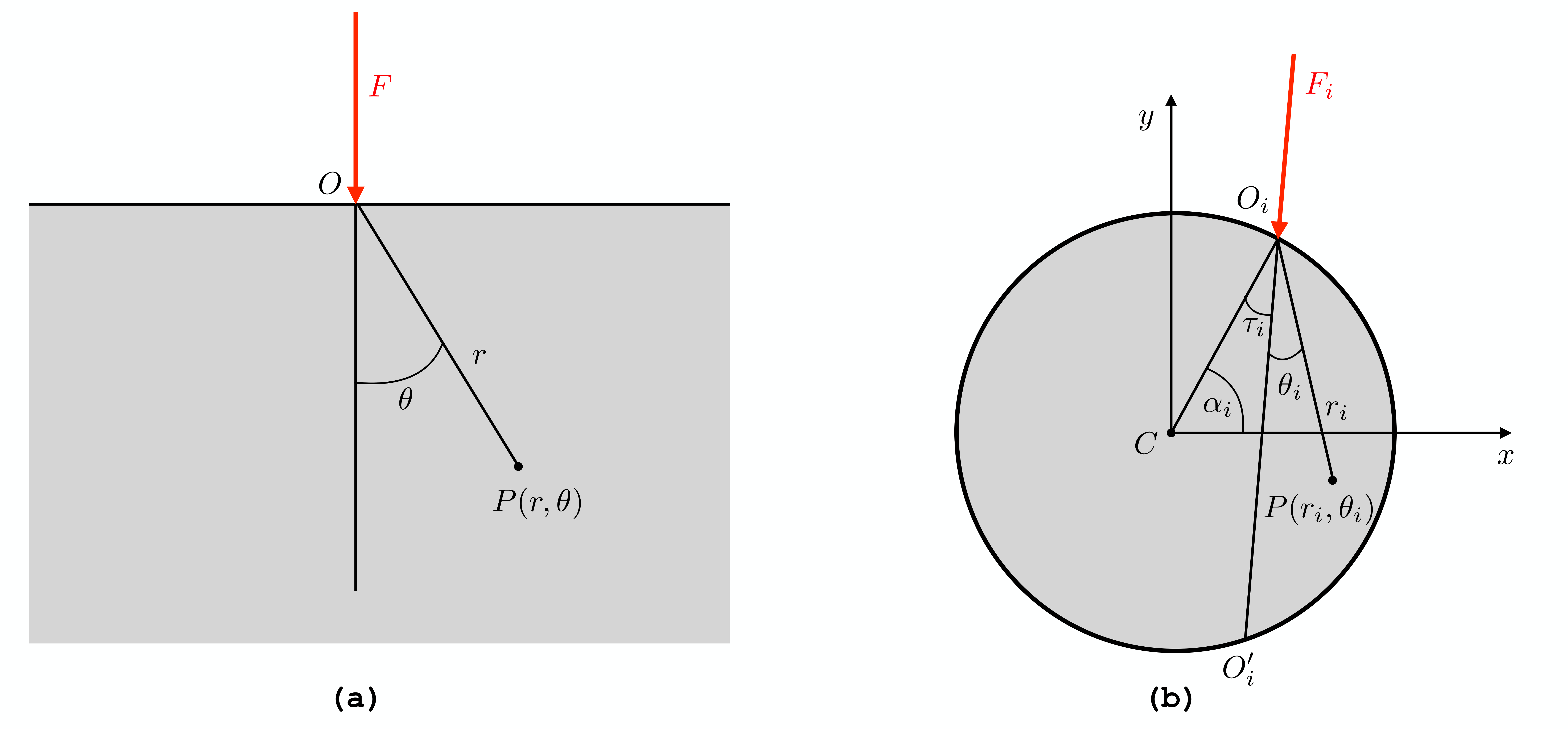}}
    \caption{(a) A semi-infinite plane with a force acting on it at a single point $O$. To compute the stress at any point $P$ it is convenient to use the polar coordinates based in the point $O$ measuring the angle $\theta$ counterclockwise from the direction of the force.  
    (b) Schematic representation of a particle. Only a single force $\{(F_i,\alpha_i, \tau_i)\}$ acting on this particle is depicted. The magnitude of this force is $F_i$. The position of the impact point $O_i$ on the boundary of the disk is encoded by the angle $\alpha_i$ measured counterclockwise from the $x$-axis. The angle $\tau_i$ indicates how much the force deviates from pointing to the center of the particle. To compute the stress caused by this force we use the polar coordinates based at the point $O_i$  measuring the angle $\theta_i$ counterclockwise from the direction of the force.   }
    \label{fig::StressTensor}
\end{figure}

To solve Equation~(\ref{eqn::FinalElastic}) we  exploit the fact that its solution is well known for a semi-infinite plane with  a single perpendicular force acting on it at some point $O$, see Fig.~\ref{fig::StressTensor}(a).  Let us consider  a polar coordinate system based in $O$ with angles measured counterclockwise from the force vector. In this coordinate system, the Airy stress function is given by $\varphi(r,\theta)= \frac{F}{\pi h}r\theta\sin\theta$, where $F$ is the magnitude of the force and $h$ is the thickness of the plane.With this in mind, the stress tensor can be written as
\begin{equation}
    \label{eqn::palnestress}
    \sigma_{rr}(r,\theta) = -\frac{2F}{\pi h}\frac{\cos\theta}{r}, \quad \sigma_{\theta\theta}(r,\theta) = 0, \quad \sigma_{r\theta}(r,\theta) = 0.
\end{equation}

Now we turn our attention to an elastic disc exposed to $M$ external forces. As indicated in Fig.~\ref{fig::StressTensor}(b), each of these forces can be encoded by a triplet $(F_i, \alpha_i, \tau_i)$. The value $F_i$ is the magnitude of the $i$-th force measured in Newtons.  The impact point $O_i$ of this force is determined by an impact angle $\alpha_i$  between the $x$-axis and the line connecting the center of the particle to  $O_i$.  Finally, $\tau_i$ is the angle between the direction of the force and the line segment $\overline{O_iC}$. The angles $\alpha_i$ and $\tau_i$ are measured counterclockwise in radians. 
The fact that Equation~(\ref{eqn::FinalElastic}) is linear makes it possible to compute the final stress $\sigma(x,y)$ by summing up the stresses $\sigma^i(x,y)$ caused by individual forces. 

Inspired by the example of the semi-infinite plane, we seek the solution for the stress due to the force $(F_i, \alpha_i, \tau_i)$ in the polar coordinates $(r_i,\theta_i)$ centered at the point $O_i$ measuring the angle $\theta_i$ counterclockwise from the direction of the force, see Fig.~\ref{fig::StressTensor}(b). In this coordinates the stress $\sigma^i$ can be written as 
\begin{equation}
    \label{eqn::istress}
    \sigma^i_{r_ir_i}(r_i, \theta_i) = -\frac{2F_i}{\pi  h}\frac{\cos\theta_i}{r_i} + \frac{F_i}{\pi R h}\cos\tau_i, \quad \sigma^i_{\theta_i\theta_i}(r_i, \theta_i) = 0, \quad \sigma^i_{r_i\theta_i}(r_i, \theta_i) = 0,
\end{equation}
where $h$ is the thickness of the particle and $R$ is its radius. The first term of $\sigma^i_{r_ir_i}(r_i, \theta_i)$ is analogous to the solution on the semi-infinite plane and the second term is added to ensure that the boundary condition (no stress except at the impact points) is satisfied. Note that each of the stress tensors $\sigma^i$ is expressed in a different polar coordinate system. To obtain the full stress tensor $\sigma(x,y)$ in  Cartesian coordinates we perform appropriate changes of coordinate systems as we sum up the individual stresses. So the total stress is
\begin{equation}
    \label{eqn::finalstress}
    \sigma(x,y) = \sum_{i=1}^M T(\theta_i(x,y)) \sigma^i(r_i(x,y), \theta_i(x,y)),
\end{equation}
where the matrix $T(\theta)$ rotates the plane by the angle $\theta$.

\section{Force Reconstruction from Photoelastic Response} 
\label{sec::reconstruction}

In the previous section, we explained how to compute the photoelastic response of a circular particle if the forces acting on it are known. Extracting the forces acting on individual particles from experimental data is a much more complicated problem.  Figure~\ref{fig::exp} is an example of an experimental image of the photoelastic patterns exhibited by a group of particles. This image is only shown to illustrate the patterns and cannot be directly used for the force reconstruction which requires monochromatic lighting of the particles. 

There are several computational packages \cite{codeGrainSolver,codePEDiscSolve} that implement the force reconstruction based on the technique pioneered in~\cite{majmudar05a}.  As explained in~\cite{daniels2017photoelastic}, these methods require two images of the same group of particles. The first image taken in unpolarized light is used to extract the positions of the particles and create a preliminary list of force-bearing contacts. This  list of forces is then refined using the photoelastic patterns visible in the second image taken in polarized monochromatic light. For each particle, an initial list of forces acting at the force-bearing contacts is guessed based on  a squared local gradient average of the pattern.  Then, an optimization algorithm is used to minimize the $L_2$ difference between the experimentally captured pattern and the pattern generated from the reconstructed forces. 

As indicated in~\cite{daniels2017photoelastic}, the positions of the particles can be computed with high accuracy using the Hough
transform. On the other hand, we are not aware of any systematic study of the accuracy of the force reconstruction.  In particular,  reconstruction of the force  acting between two particles can considerably vary depending on which of the two particles is used for the reconstruction~\cite{daniels2017photoelastic}. Therefore,  we consider the primary problem to be force reconstruction from the photoelastic response for the individual particles. 

In this paper, we reconstruct the forces $\{(F_i,\alpha_i, \theta_i)\}_{i=1}^M$ acting on a  particle by exploiting the natural constraints due to  geometry and physical properties of the particles typically used in experiments.  We summarize these constraints in the rest of this section.

In principle, there could be a large number of forces acting on a single particle but this is not the case if the sizes of particles in the ensemble do not vary too much, see Fig.~\ref{fig::exp}. In this case, the number of forces acting on a  particle may be assumed to be less than or equal to six. To avoid technicalities and demonstrate the ideas, we concentrate on bi-disperse particles. By simple modifications indicated in Section~\ref{sec::conclusions}, our method can be expanded to polydisperse materials. In the case of bi-disperse particles the impact points at which the forces act on the particle cannot be too close to each other. To be more precise, the angles $\alpha_i$ encoding positions of the impact points can be assumed to satisfy 
\begin{equation}
    |\alpha_i - \alpha_j| \geq \frac{\pi}{6}, \text{ if } i\neq j.
\end{equation}

The photoelastic theory presented in Section~\ref{sec::photoelasticity} is only valid if the particles are in mechanical equilibrium. Thus, to ensure that the reconstruction works properly,  particles have to be allowed to  equilibrate before collecting the data.  The fact that the particles are in a mechanical equilibrium imposes force balance and torque balance constraints on each particle. Moreover, the force balance on a particle implies that $M\geq 2$.

Finally, the friction coefficient $\mu$ of the particles is typically smaller than $1$ and so  the angles $\tau_i$ satisfy 
\begin{equation}
-\frac{\pi}{2} \leq \tau_i \leq \frac{\pi}{2}.    
\end{equation}
For example, the friction coefficient of the particles shown in Fig.~\ref{fig::exp} is $\mu=0.8$ as reported in~\cite{wang2018microscopic}. 

\section{Machine Learning Approach to Force Reconstruction}
\label{sec::MLReconstruction}

In this section, we set up a machine learning framework for reconstructing the forces acting on a particle from its photoelastic response. We use convolutional neural networks (CNNs) in conjunction with transfer learning~\cite{pan2009survey, zhuang2020comprehensive, yosinski2014transferable, sharif2014cnn}.  While CNNs are  well suited for the recognition of spatial patterns in digital images~\cite{lecun2015deep, krizhevsky2012imagenet, simonyan2015a}, transfer learning yields  outstanding results for a wide range of applications such as medical image analysis, speech emotion recognition, and quantum phase transition identification \cite{shin2016deep, tajbakhsh2016convolutional, deng2014autoencoder, huembeli2018identifying}.   We utilize inductive transfer learning, in which a model originally trained to perform some (unrelated) task is retrained to carry out a new  task. We show that the previous knowledge encoded in the model is transferred to the new task using a relatively small training set. Furthermore,  in Section~\ref{sec::results}, we demonstrate that the models trained on the synthetic data are extremely likely to  transfer well to the experimental data even on much smaller training sets.

\subsection{Pipeline Architecture}
\label{sec::pipeline}

A single CNN model to reconstruct the forces acting on a particle  would have to be rather complex and contain many layers. Therefore, a  large training set would be necessary to prevent overfitting.  We avoid this problem  by dividing the reconstruction of the force list $\{(F_i,\alpha_i,\tau_i)\}_{i=1}^M$  into four smaller tasks. 

The first task is to infer the number $M$ of the forces acting on the particle.  For this purpose, we use a single CNN model. The reconstruction is then  further divided into three additional tasks: inferring the magnitudes $F_i$, reconstructing the angles $\alpha_i$, and finally estimating $\tau_i$.   As explained in Section~\ref{sec::reconstruction}, we assume that  $2\leq M \leq 6$. Thus, for each of the 3 additional tasks we implement a collection of five independent CNN models. In other words, we use one model for each task and every possible value of $M$.

\subsection{Models}
\label{sec::transfer}

We experimented with transfer learning for the individual tasks by using three models that show state-of-the-art performance on the benchmark data sets. In particular we considered the VGG19, InceptionResNetV2, and Xception models, pre-trained on the ImageNet data set ~\cite{simonyan2015a, szegedy2017inception, chollet2017xception}. The ImageNet data set is a benchmark data set which contains $3.9$ million labeled images of real-world objects divided into $1,000$ distinct classes. The model VGG19 developed by \cite{simonyan2014very} is an example of a traditional deep convolutional neural network, whose main power lies in its depth. On the other hand, InceptionResNetV2 is built upon the Inception architecture and further adds the residual connections to boost the performance as outlined in \cite{szegedy2017inception}. Finally, the Xception model is again based on the Inception architecture but replaces the Inception modules with depthwise separable convolutions \cite{chollet2017xception}. 

To train the models on our data set, we remove the last fully connected layer for the Xception and InceptionResNetV2 models and the last three fully connected layers for the VGG19 model. Instead, we placed a global averaging layer followed by two dense layers with a $50\%$ drop out in between them for the  Xception and InceptionResNetV2 models. For the VGG19 model, we place a flattening layer again followed by two dense layers with a $50\%$ drop out in between them. The first dense layer for all models is followed by ReLu activation. Depending on the task of the model we use either SoftMax or linear activation for the last dense layer, which is essentially the output layer. 

\section{Training of the Neural Networks}
\label{sec::training}
\begin{figure}
    \centering
    \includegraphics[width = \textwidth]{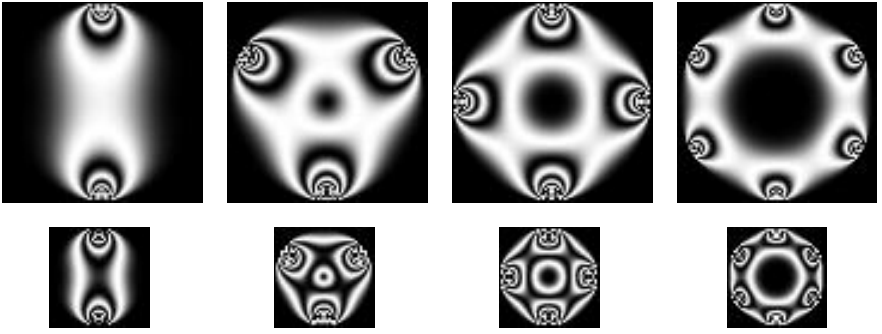}
    \caption{Top row: Computer-generated photoelastic responses on a particle with radius $r=0.008$ for a variety of different forces acting on the particle. Bottom row: Computer-generated photoelastic responses on a particle with radius $r=0.004$ for the same forces as in the top row. }
    \label{fig::small_vs_large}
\end{figure}

We train and test our models on the synthetic data generated using the algorithm described in  Section~\ref{sec::photoelasticity}. Training the models on synthetic data can be desirable when the labeled experimental data is scarce.  This strategy was already successfully used for the tasks of object detection, optical flow estimation, and scene understanding   \cite{tremblay2018training, handa2015scenenet, mayer2016large}. To show that the models pre-trained on the synthetic data are likely to transfer well to experimental data we use two different sizes of particles.  We first train the models on particles with radius $r = 0.008$ and then take these pre-trained models and retrain them on particles with $r = 0.004$ to simulate performance on the experimental data. Figure~\ref{fig::small_vs_large} shows that changing the radius of the particles alters the exhibited photoelastic patterns considerably. Differences are typically larger than between the synthetic and experimental data for the particles of the same size. Thus a successful transfer between the particles of the different sizes indicates the potential of this method for the experimental data.  Before discussing the training of different models we first summarize the protocol for generating the data sets used for training, validation, and testing purposes. 

\subsection{Generating the Data Sets}
As explained in Section~\ref{sec::photoelasticity}, the photoelastic pattern on a particle can be computed from the list of forces $\{(F_i, \alpha_i, \tau_i)\}_{i=1}^M$ acting on it. To decrease the bias of the model the force lists in the training set must form a representative sample~\cite{chollet2017deep}. To ensure that the generated force lists are realistic we enforce the constraints presented in Section~\ref{sec::reconstruction}. We require that forces and torques are balanced and the normal component of the force is larger than its tangential component, i.e. $|\tau_i| \leq \pi/2$. For bi-disperse systems, considered in this paper, we assume that $2\leq M \leq 6$ and the angles of distinct impact points satisfy $|\alpha_i-\alpha_j| \geq \pi/6$.

To create a diverse training set with realistic forces we use the following procedure. For each value of $M$ we create the same number of force lists.  To construct a force list $\{(F_i, \alpha_i, \tau_i)\}_{i=1}^M$  we start by randomly generating the first $M-1$ forces. The last force, described by  $(F_M, \alpha_M, \tau_M)$,  is then added to achieve force and torque balance.  The angles 
\begin{equation} 
\label{eqn::alpha}
\alpha_i \sim {\mathcal U}\left(\frac{2\pi (i-1)}{M}+\frac{\pi}{12}, \frac{2\pi i}{M}-\frac{\pi}{12}\right), \quad 1 \leq i \leq M-1,
\end{equation}
are drawn from uniform distributions.
The magnitudes of the first $M-1$ forces are drawn from a normal distribution: 
\begin{equation} 
F_i  \sim {\mathcal N}\left(F, \left(\frac{F}{5}\right)^2\right),
\end{equation}
where $F$ is sampled from the exponential distribution  the rate parameter  $1/2$ truncated to the interval $[0.01,0.9]$. This distribution closely resembles the distribution obtained experimentally  in~\cite{dijksman2018characterizing}. Finally, each angle $\tau_i$ is taken from the normal distribution
\begin{equation} 
 \tau_i  \sim {\mathcal N}\left(0, \frac{\pi}{12}\right).
\end{equation}
After producing a force list we check that all the assumed conditions from Section \ref{sec::reconstruction} are satisfied.  If these conditions are not satisfied, then we generate a new force list and discard the old.

For each accepted force list we produce a photoelastic pattern generated by this list. The resolution of the digital images depicting the patterns is $0.00019$ meters per pixel for both small and larger particles.  The data set for the particle with radius $r = 0.008$ contains $24000$ distinct force lists ($4800$ for each possible value of $M$) accompanied by the corresponding photoelastic patterns. For the smaller particle, we generate a data set consisting of  $7400$ force lists ($1480$ for each possible value of $M$).

We split the generated data sets, for each value of $M$, randomly into the training, validation, and test sets as follows. For the large particle, we use $3200$ samples for the training set, $800$ for the validation set and the test set. For the small particle, we create a sequence of three independent training sets with sizes $64, 160$, $640$ and three independent validation sets with sizes $16, 40, 160$. The common test set contains $400$ samples.

Separately, for each of the sets, we perform the following data augmentation to boost the generalization power of the models. 
We randomly rotate each sample  $5$ times using random angles drawn from the uniform distribution $\mathcal{U}(0, 2\pi)$. To achieve a uniform size of the image, we resize them to $128 \times 128$ pixels using the nearest neighbor interpolation. Finally, to ensure a natural look of the rotated images we apply Gaussian blur with standard deviation $\sigma = 1$~\cite{bishop1995training, goodfellow2016deep}.

\subsection{Training the  Models }

We recall that our pipeline contains four units. The first classification unit infers the number of forces acting on the particle while the other three regression units reconstruct the individual components of the force list. As suggested in~\cite{chollet2017deep}, we use the mean absolute error as a loss function for training the regression units. For the classification unit, we utilize one hot encoded vector to describe the probabilities that the object belongs to a particular class and employ the categorical cross-entropy as a loss function. For the optimization algorithm, we chose to use Adam optimizer \cite{chilimbi2014project}, which is the default algorithm used to train all the models \cite{simonyan2015a, szegedy2017inception, chollet2017xception}.

We start by training only the last fully connected block while freezing the rest of the model. This warm-up period is important because the weights for the new output block are initialized randomly. Skipping this step could corrupt the weights of the original pre-trained model, which are supposed to provide useful representations of the data \cite{pan2009survey}. To further decrease the chance of breaking the pre-trained model during its fine-tuning when all the weights are allowed to change,  we decrease the learning rate from the standard value $10^{-3}$ to $10^{-6}$ as suggested in \cite{yosinski2014transferable}. We set the number of epochs to $200$ for the warm-up and $100$ iterations for the main training. We further implement early stopping and best model selection. For the early stopping, we set the patience to $20$, which means that if the error does not decrease for more than $20$ iterations, the training is finished. The best model selection is implemented by only saving the best-performing model. 

\subsection{Training on a Smaller Particle}

For the small particles, instead of retraining the models from scratch, we reuse the models trained on the larger-sized particles.  To alleviate the computational burden, for each task we only consider the model with the best performance on the larger particle.  As explained above, we use three pairs of train and validation sets with increasing sizes and one common test set. This is done to test the power of the transfer learning \cite{pan2009survey, zhuang2020comprehensive}. In particular, we are interested in the size of the data set needed to achieve reasonable accuracy.  By pairing each train set with a separate validation set, we prevent any possible data leakage. The common test set is used to measure the final accuracy of the whole model pipeline.

We train the models by leaving all the structural components intact and as before using the learning rate  $10^{-6}$ because we are just fine-tuning the models. The tasks for the large and small particles are relatively similar, so we neither reprogram the top layer nor perform any warm-up~\cite{yosinski2014transferable}. 

\section{Results}
\label{sec::results}
In this section, we demonstrate our results for reconstructing the forces acting on a particle from its photoelastic response.  We start by comparing and contrasting the performance of the different models for the particle with radius $r = 0.008$. 

\begin{table}
\centering
\begin{tabular}{|c|c|c|c|}
\hline
   Model      & VGG19 & Xception & InceptionResNetV2   \\ \hline
Accuracy & 0.9999 & 0.9835 & 0.9975                \\ \hline
\end{tabular}
\caption{Accuracy of different models used for detecting  the number of forces acting on a particle with radius $r=0.008$.}
\label{tab::accuracy}
\end{table}

\begin{figure}
    \centering
    \includegraphics[width = 50mm]{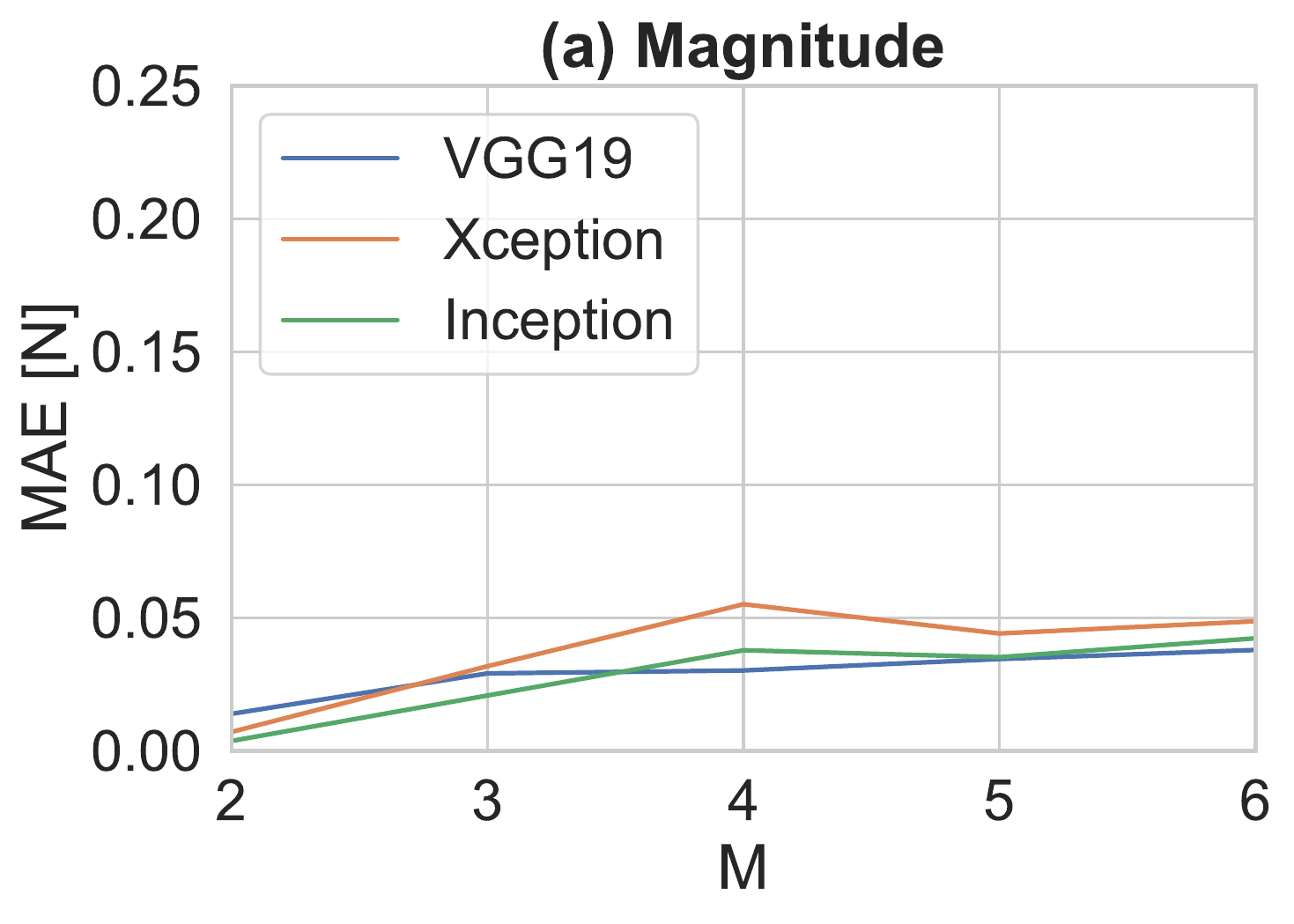}
    \includegraphics[width = 50mm]{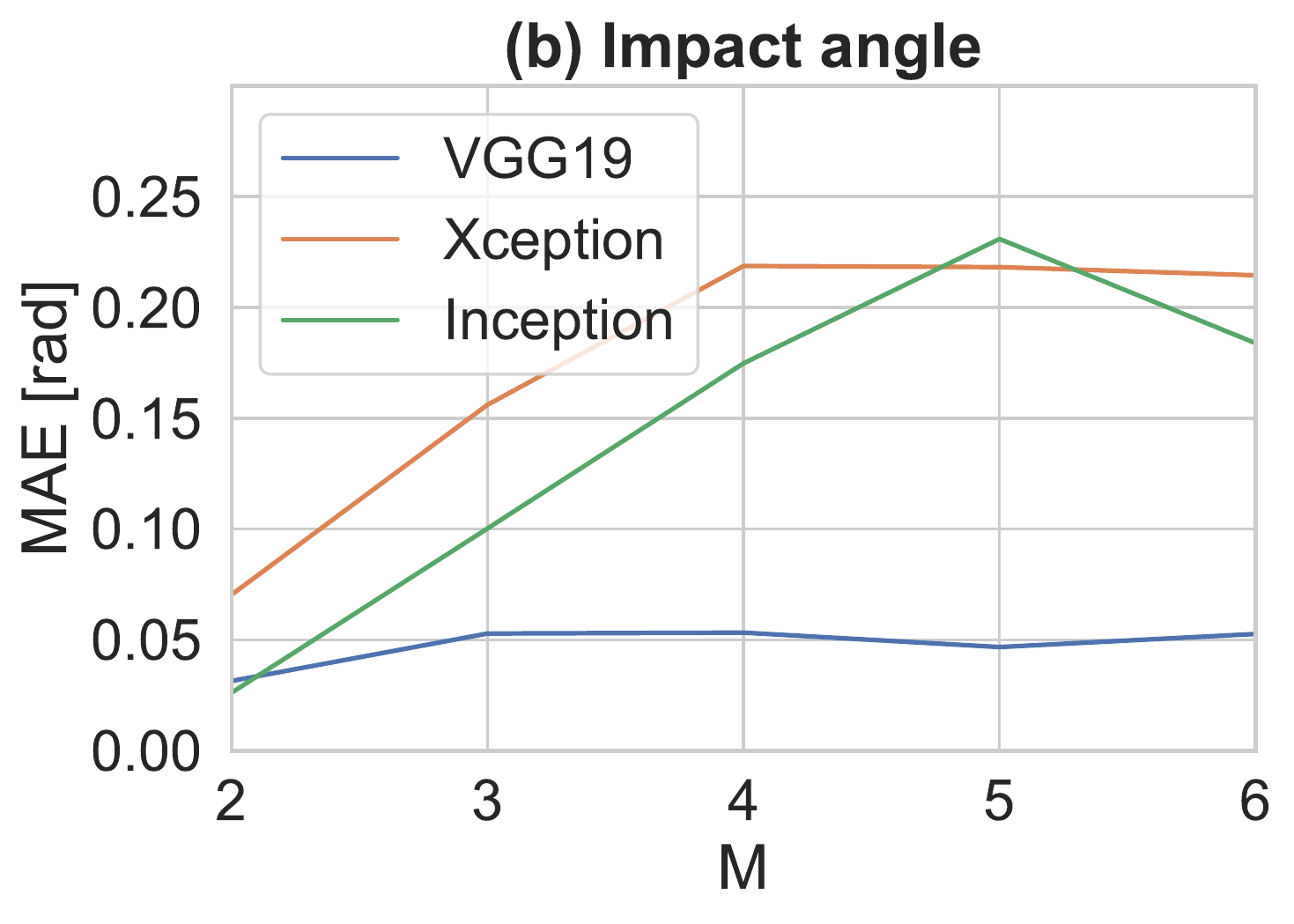}
    \includegraphics[width = 50mm]{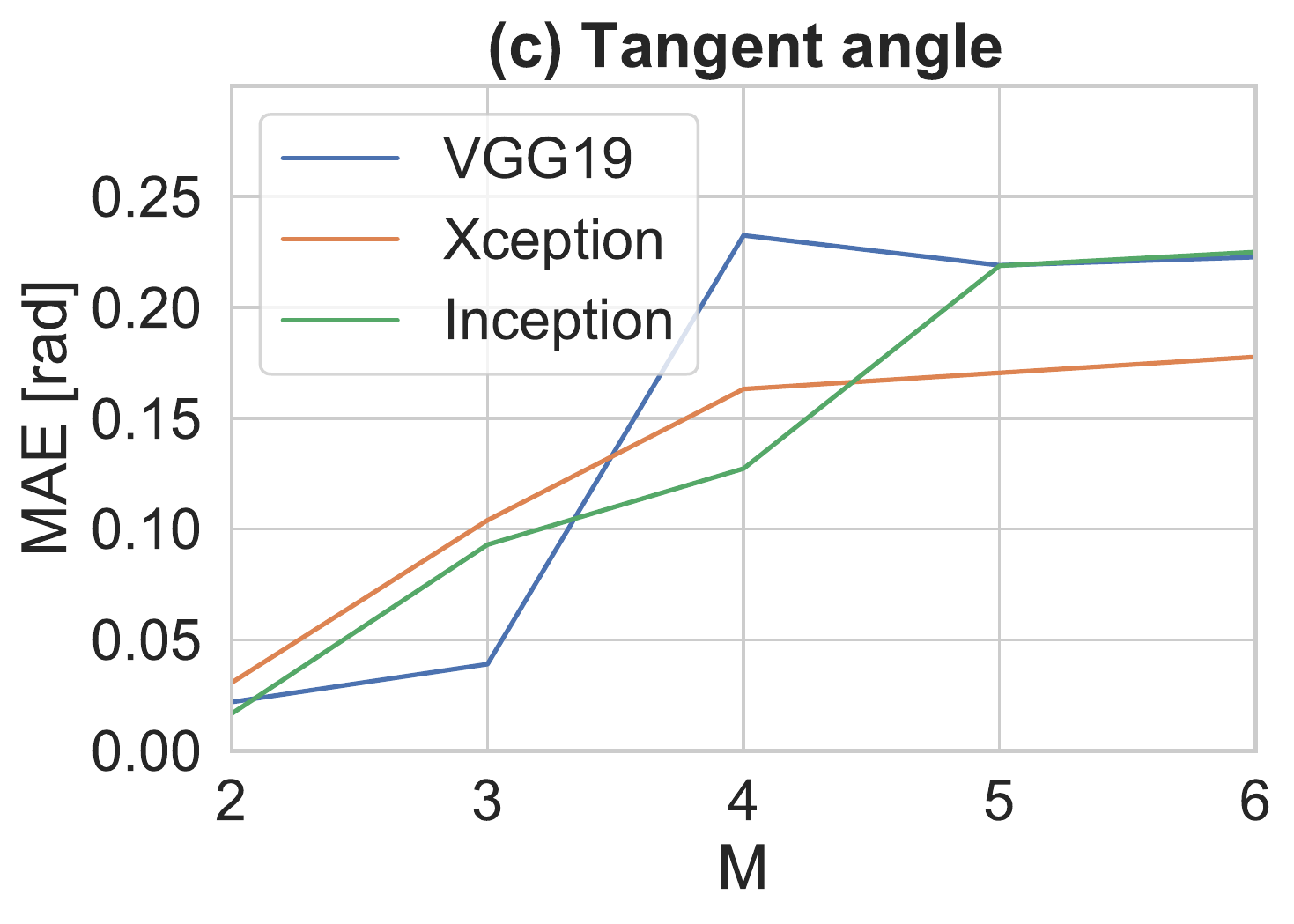}
    \caption{Mean absolute error (MAE) of the (a) force magnitude, (b) impact angle  and (c) tangent angle for the 
    particle with radius $r = 0.008$. Each panel shows the errors for  VGG19, Xception, and InceptionResNetV2 models. We report the errors for each model and different  numbers of forces, $M$, acting on a particle.}
    \label{fig::large-mag-alpha-tau}
\end{figure}

Table~\ref{tab::accuracy} shows that the accuracy of detecting the number of forces acting on a particle is very high for all the models. Given the same time to train and similar training procedures, the VGG19 model marginally outperforms the other two. It is not only the high accuracy that makes our approach to detecting the number of forces appealing. Unlike the classical algorithms for detecting force-bearing contacts given in \cite{daniels2017photoelastic}, our method does not require any adjustment of the parameters from the user. Since the misclassification is very rare, we test the subsequent models in the pipeline assuming that the number of forces acting on the particle is known.

Now we consider the the accuracy of the reconstructed force lists $\{(F_i,\alpha_i, \tau_i)\}_{i=1}^M$. Figure~\ref{fig::large-mag-alpha-tau} shows the mean absolute error (MAE) of the force magnitude, impact angle, and tangent angle for VGG19, Xception, and InceptionResNetV2 models. We report the errors separately for different values of $M$. The general trend is that the error increases with $M$. This could be caused by the fact that the patterns tend to be more complex for large values of $M$. The errors in the magnitude stay reasonably small for all considered models and values of $M$. This is not the case for the impact angle. The VGG19 shows much better performance than the other two models. The mean absolute error of the VGG19 model is below three degrees for all values of $M$. This means that the mean error in detecting the impact point is around two pixels. The accuracy of the tangential angle reconstruction is much worse for all three models. For $M>3$ it can reach $10$ degrees. We expect that this is caused by the fact that changes of this angle have a less dramatic influence on the photoelastic pattern than the magnitude and impact angle. 

\begin{figure}
    \centering
    \includegraphics[width = 50mm]{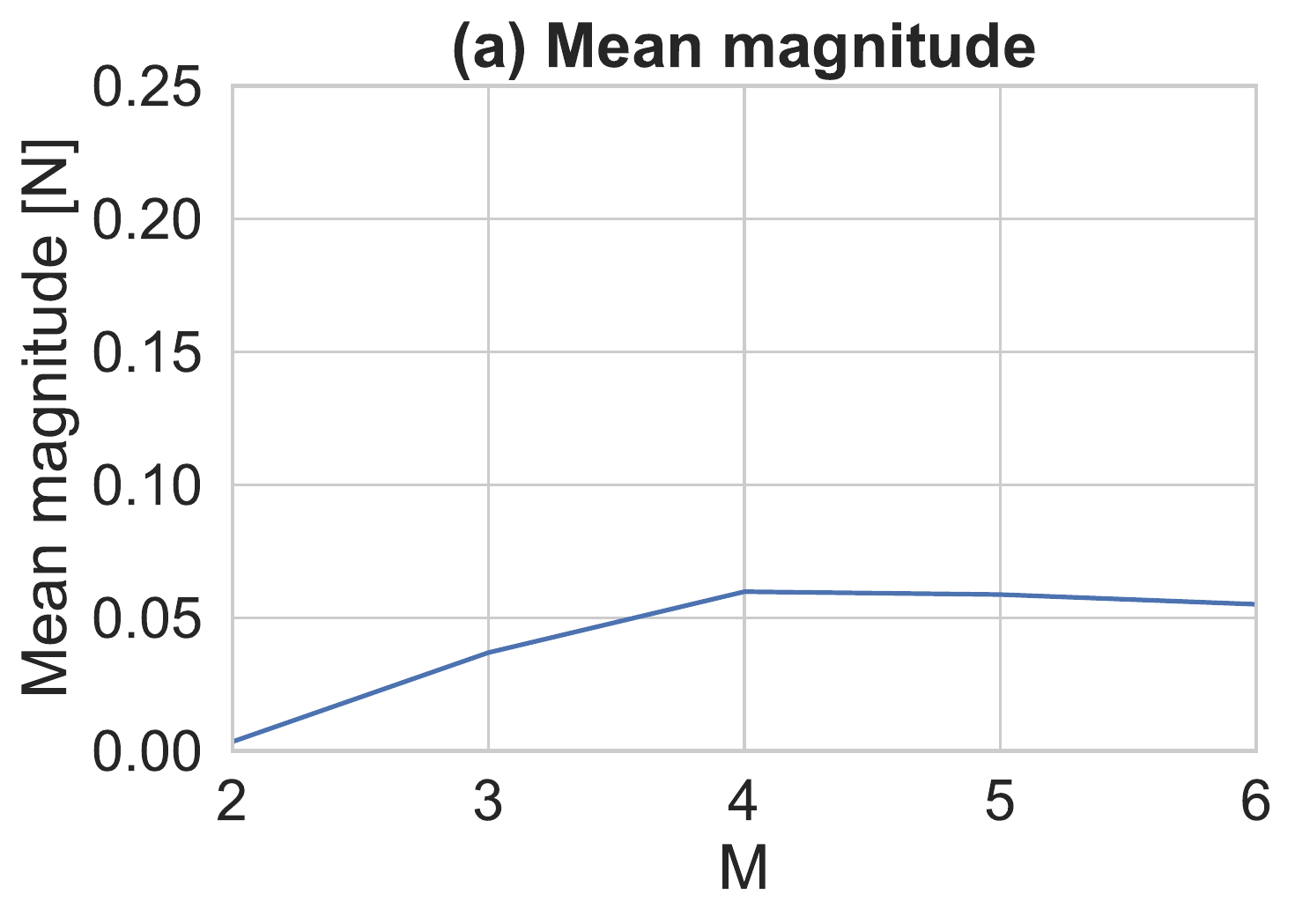}
    \includegraphics[width = 50mm]{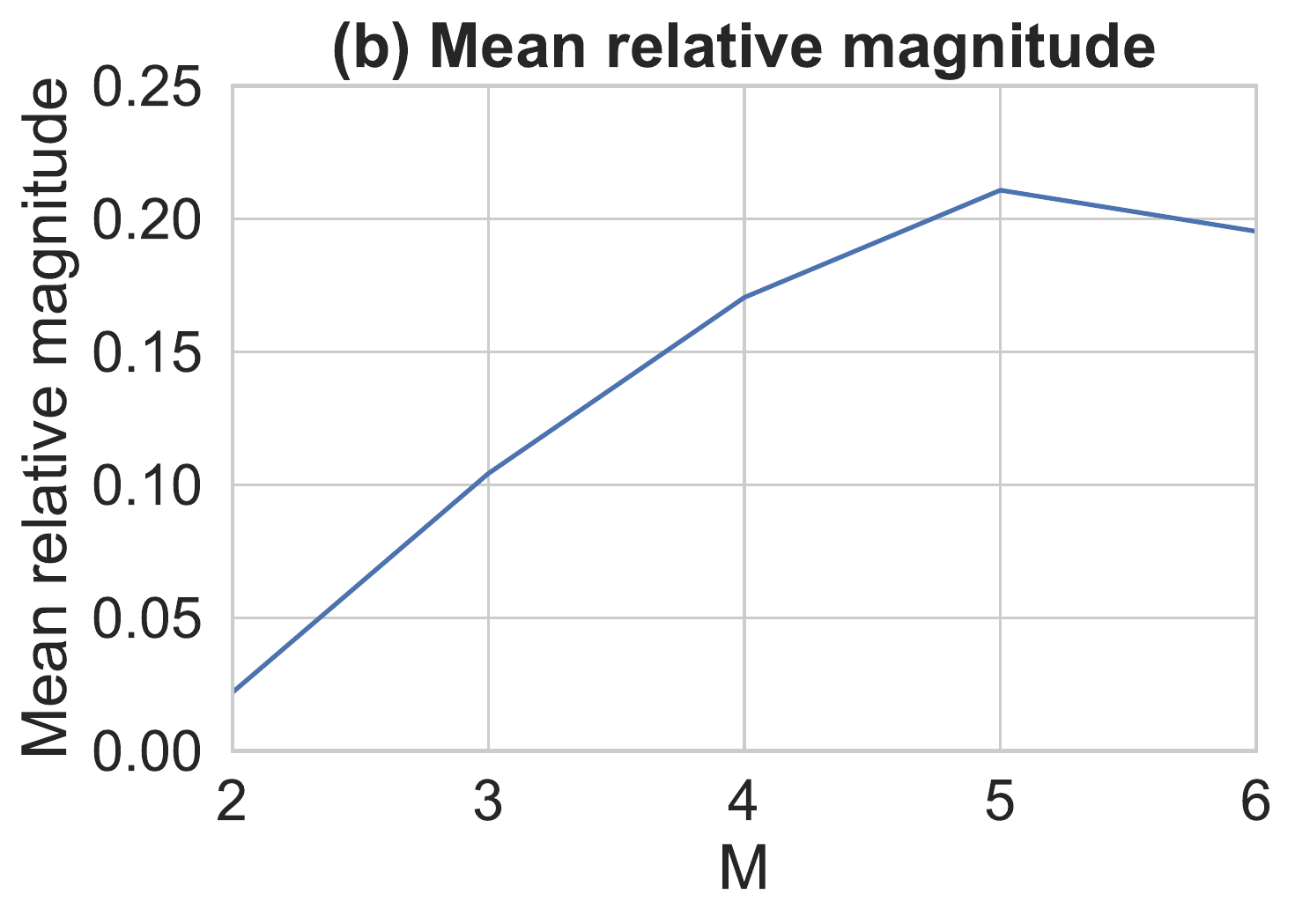}
    \caption{(a) Mean magnitude of the sum of the forces acting on a particle with radius $r = 0.008$. (b) Mean magnitude, shown in part (a), normalized by the average magnitude of the forces acting on the particle. We report the errors separately  for  different  numbers  of  forces, M,  acting  on  the  particle.}
    \label{fig::resulting-force}
\end{figure}

We recall that the forces acting on the particle are required to be balanced. Hence, another more global way of accessing the quality of the reconstruction on a particle is to sum up the reconstructed forces. If the reconstruction works well the  magnitude of the resulting vector $\mathbf v$ should be small. Figure~\ref{fig::resulting-force} shows that this is indeed the case. However, one can consider the magnitude  of $\mathbf v$  normalized by the average magnitude of the forces acting on the particle. As shown in Fig.~\ref{fig::resulting-force}, the mean of this quantity can be as large as $20\%$ of the average magnitude. We further investigated this issue and  found out that poor reconstruction tends to occur if the magnitudes of forces are relatively small magnitude. The classical methods exhibit the same problem  and for this reason the small forces are often disregarded~\cite{dijksman2018characterizing}.

\begin{figure}
\centering
    \includegraphics[width = 50mm]{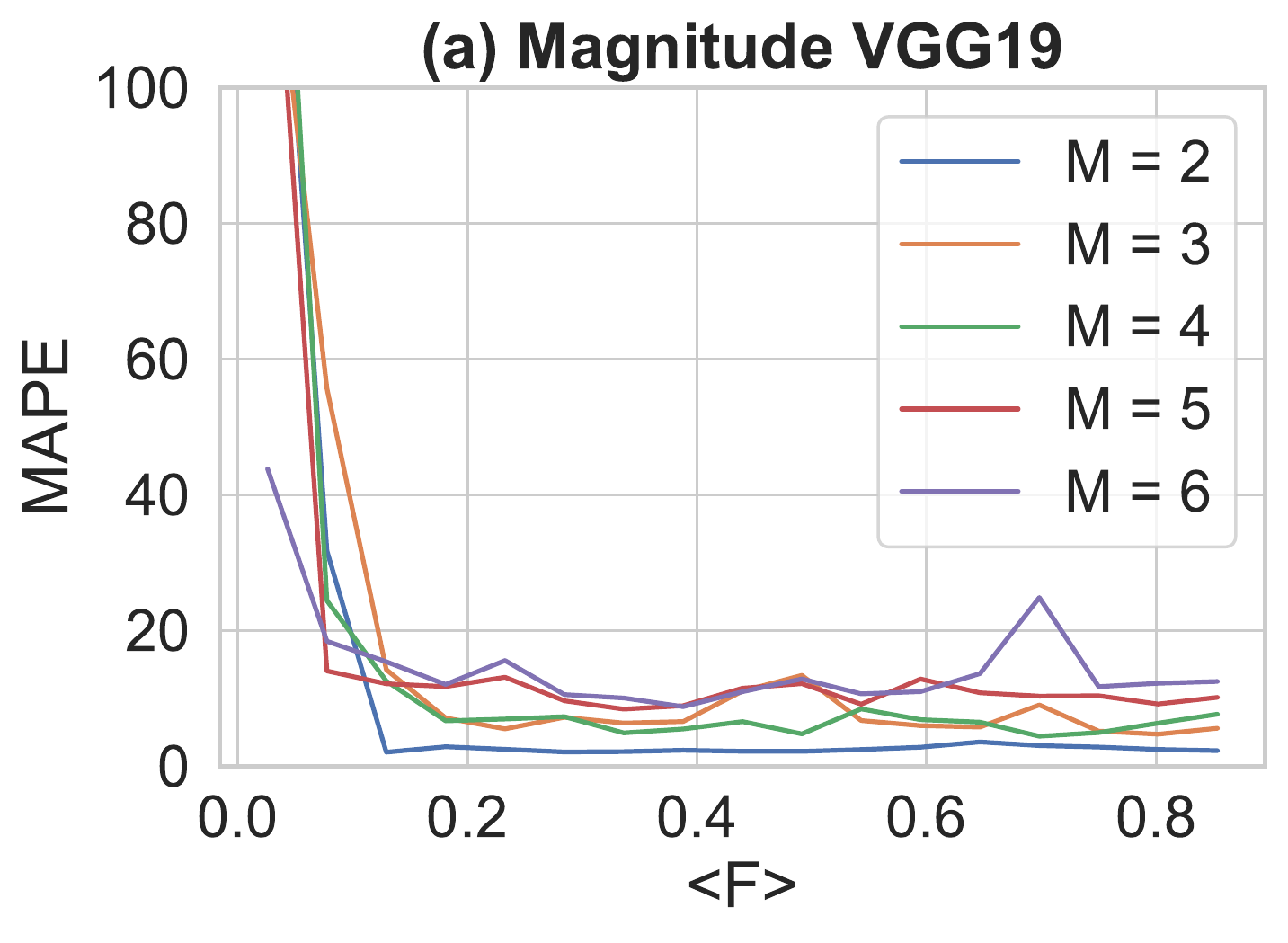}
    \includegraphics[width = 50mm]{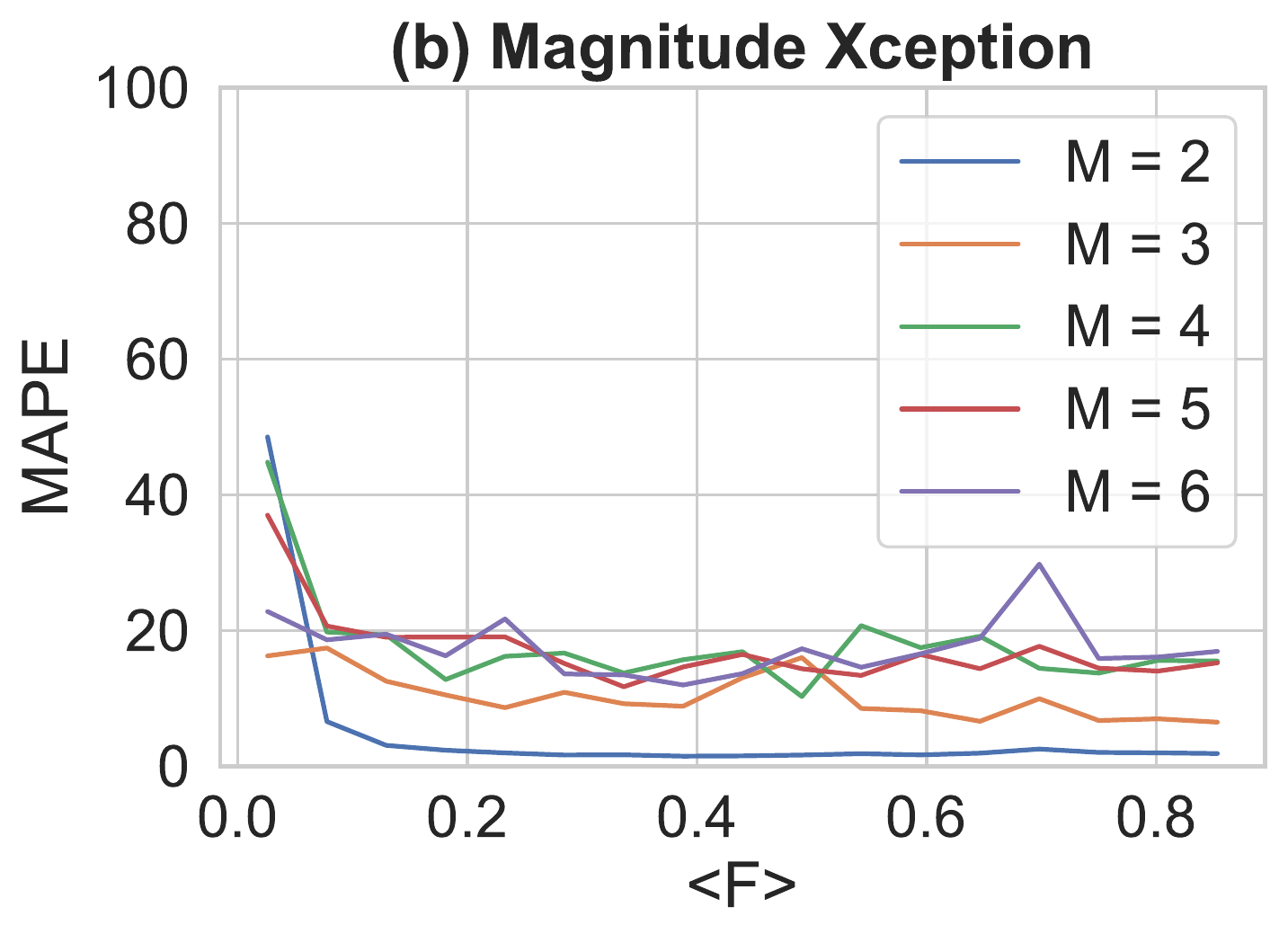}
    \includegraphics[width = 50mm]{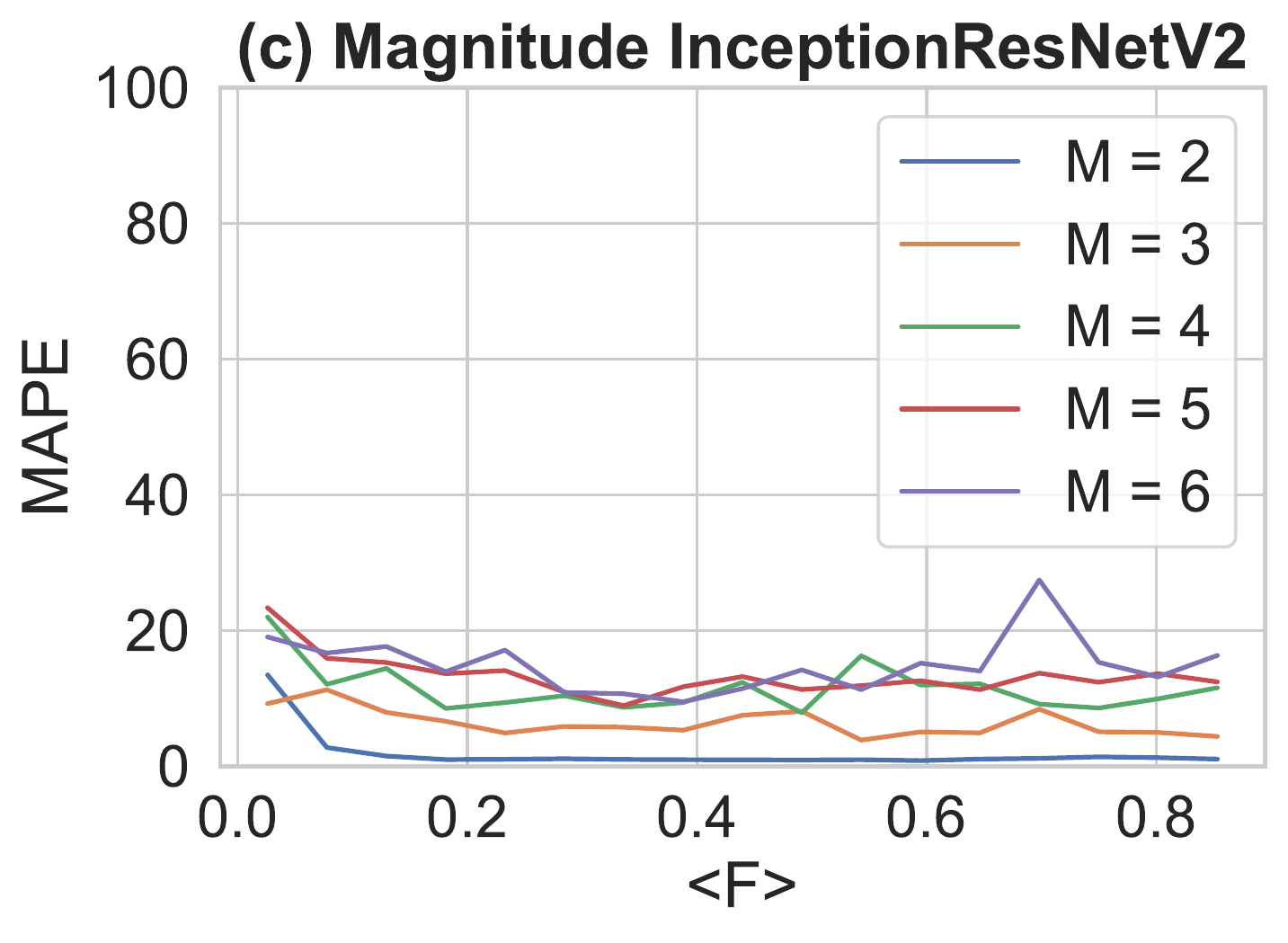}
\caption{Dependence of the mean absolute percentage error (MAPE) on the average magnitude of the forces, $\langle F \rangle$, acting on the particle with radius $r = 0.008$. We report the errors separately for different numbers of forces, $M$, acting on a particle.  The errors for different models are shown as follows  (a) VGG19, (b) IncpetionResNetV2 and (c) Xception.} 
\label{fig::mags}
\end{figure}

To better understand the relative error of the force magnitude reconstruction  we analyze its dependence on the average magnitude of the forces $\langle F \rangle$ acting on the particle. Figure~\ref{fig::mags} show how the  mean absolute percentage error (MAPE) changes with $\langle F \rangle$. As expected the relative error is typically larger for small values of $\langle F \rangle$. Figure~\ref{fig::mags} also shows that the InceptionResNetV2 model has the best performance. 

Finally, we investigate how  our models transfer to the particle with radius $r=0.004$. Figure~\ref{fig::small_vs_large} shows that the same forces produce considerably different patterns but the nature of the patterns remains similar. Hence it is reasonable to expect that the models transfer well.   To demonstrate this we only consider the best performing model for each task: VGG19 for detecting the number of forces and impact angle, InceptionResNetV2 for magnitudes, and Xception for tangent angle.  Individual models are retrain on independent data sets with increasing sizes $S=320$, $S=800$ and $S=3200$. 

\begin{table}
\centering
\begin{tabular}{|c|c|c|c|}
\hline
 Size of the data set        & 320  & 800 & 3200 \\ \hline
Accuracy & 0.9828 & 0.9850   & 0.9884            \\ \hline
\end{tabular}
\caption{Accuracy of detecting  the number of forces acting on a particle with radius  $r=0.004$ achieved by the VGG19 model trained on data sets  with different sizes.}
\label{tab::small_accuracy}
\end{table}

\begin{figure}
    \centering
    \includegraphics[width = 50mm]{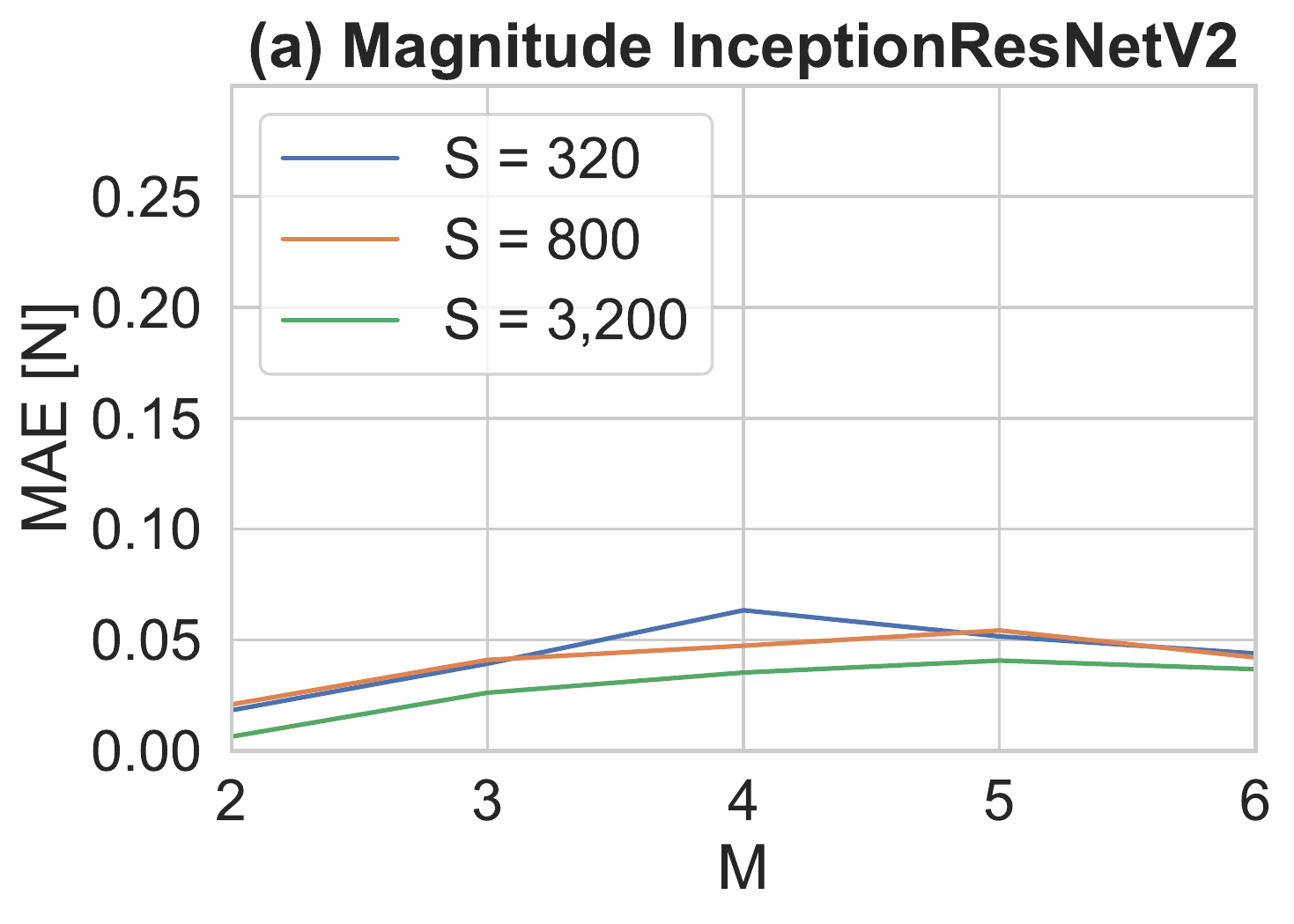}
    \includegraphics[width = 50mm]{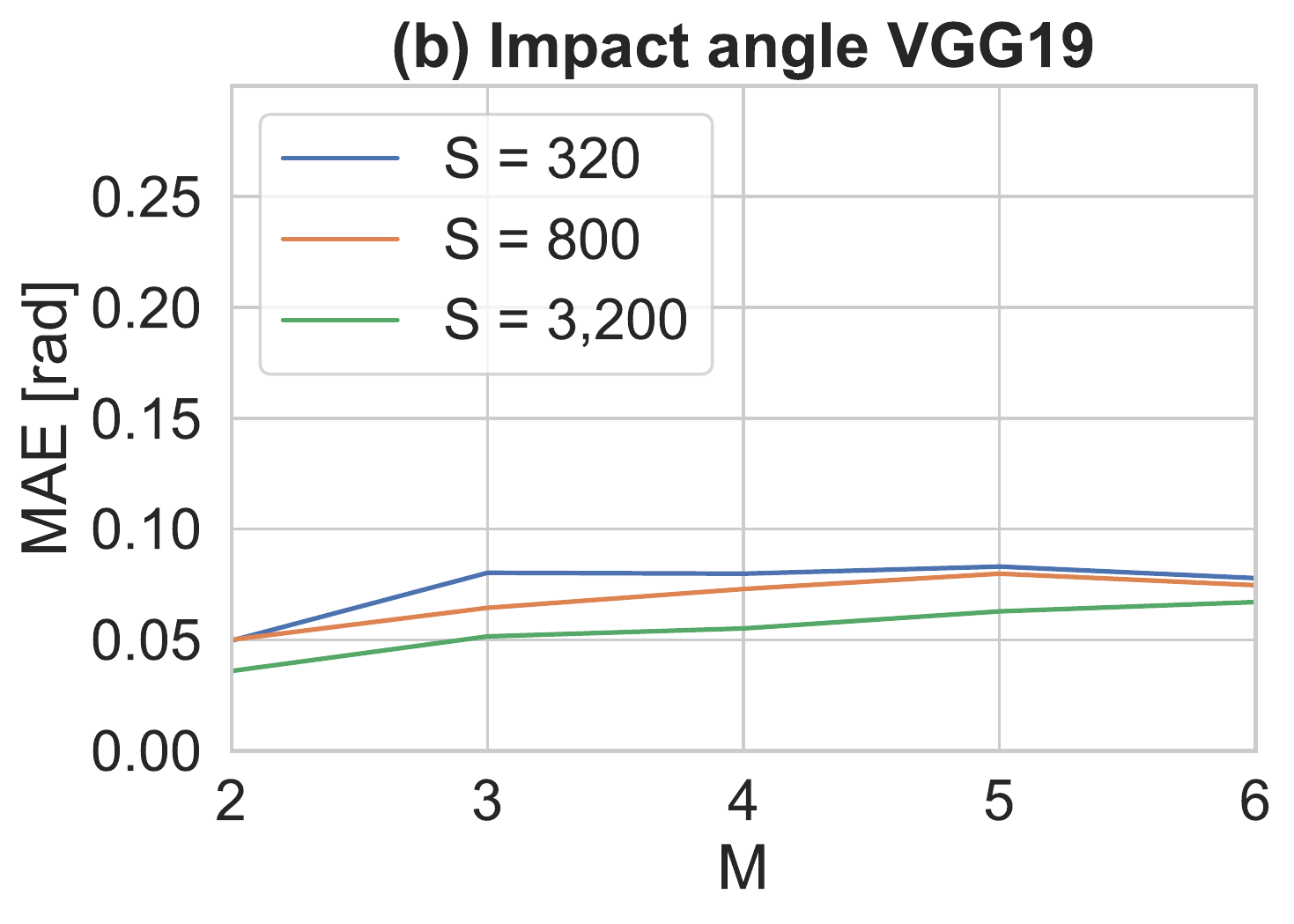}
    \includegraphics[width = 50mm]{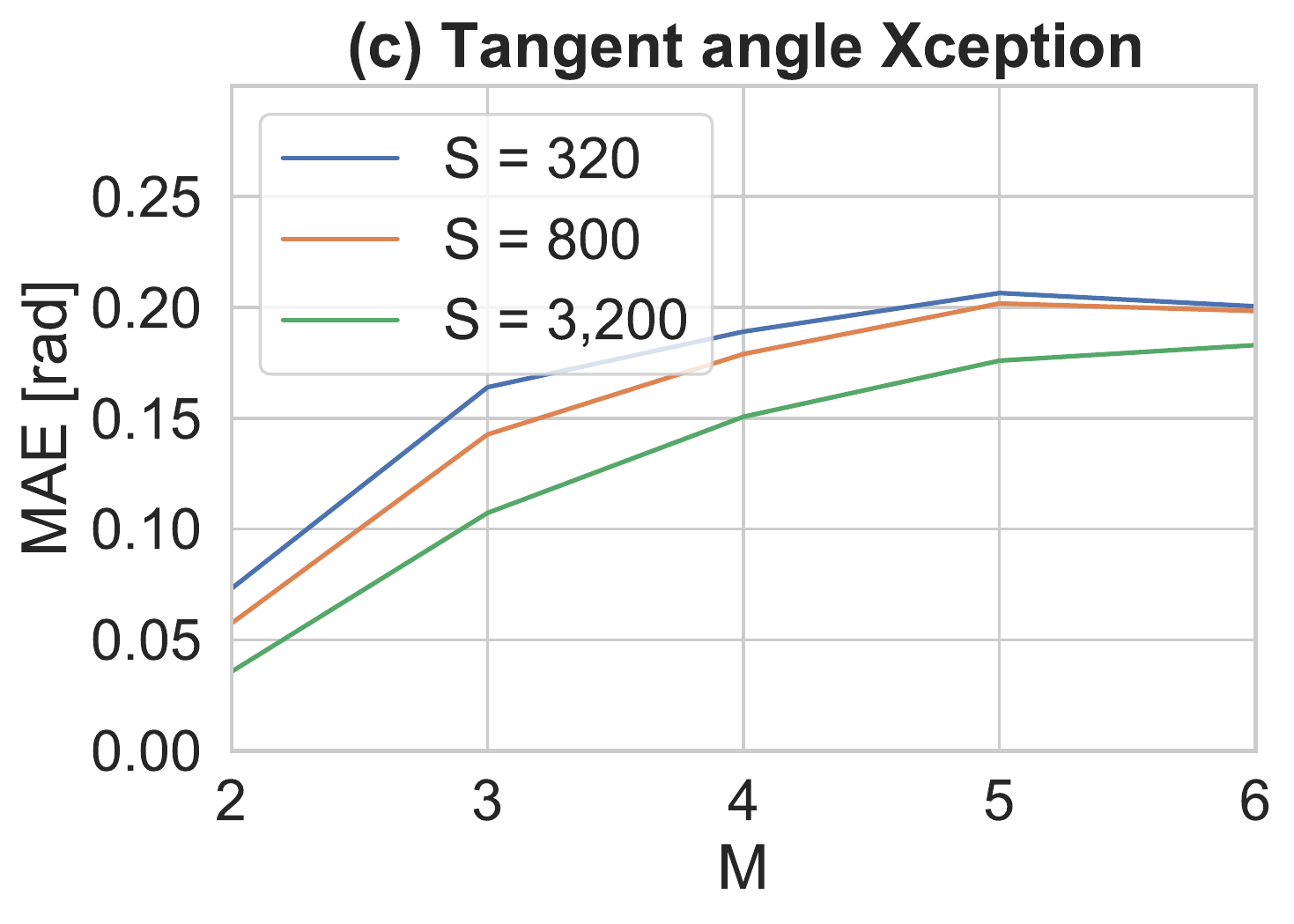}
    \caption{ Mean absolute error (MAE) for the particle with radius $r=0.004$ achieved by using training sets with different size, $S$. 
    MAE for the  (a) force magnitude obtained by InceptionResNetV2, (b) impact angle obtained by VGG19 and (c) tangent angle obtained by Xception. We report the errors for different numbers of forces, $M$, acting on a particle.}
    \label{fig::small_angles}
\end{figure}

Table~\ref{tab::small_accuracy} shows that high accuracy of detecting the number of forces is achieved on small data sets. Figure~\ref{fig::small_angles} indicates that accuracy comparable to the accuracy obtained on the large particle can be achieved on a much smaller data set. Similar analysis as performed for the larger particle shows that the relative error is larger if the average magnitude of the forces acting on the particle is small.

Differences between the patterns exhibited by the large and small particles exposed to the same forces, shown in Fig.~\ref{fig::small_vs_large}, are larger than differences between the experimental and computer generated patterns. This indicates that the complexity of transferring the models from the synthetic data to experimental data should not be larger than the complexity of transferring between particles of different sizes. Therefore, we assume that our models could be transferred easily to the experimental images.

\section{Conclusions}
\label{sec::conclusions}

In this paper, we showed the strong potential of using CNNs for reconstructing forces acting between particles in photoelastic granular material.  We compared the performance of the VGG19, Xcpetion, and InceptionResNetV2 models. We conclude that some models work better than others for reconstructing particular components of the forces. However, to achieve the high accuracy of our predictions we had to train the models on a large labeled data set. Unfortunately, it is almost impossible to produce a data set of this size experimentally. Thus, we demonstrated that training on a large experimental data set could be avoided by first pretraining the models on a  synthetic data set and then fine-tuning them on a much smaller set produced experimentally. We demonstrate this by showing that the models trained on a given particle transfer excellently to a  smaller particle and achieve high accuracy on a rather small training set.

For simplicity, we only considered bi-disperse granular materials. This assumption reduces the number of possible forces acting on an individual particle to six. However, in a particular experiment, the number of forces acting on a particle can be larger and the range of these forces can differ from our choice.  In that case our model, available at~\cite{models2020code}, should be retrained using an appropriate synthetic data set before fine-tuning it to the experimental data. To ensure that the number of forces acting on each particle is detected properly, we suggest using a slightly broader range of forces than the range expected in the experiments. 

One of the main advantages of using CNN is that after the models are trained the force reconstruction becomes extremely fast. Therefore, this approach can be used to analyze the time evolution of granular ensembles consisting of a large number of particles.

Finally, to expand our approach to polydisperse materials, used in a variety of experiments~\cite{majmudar05a, wang2018microscopic,dijksman2018characterizing,tordesillas2012transition}, we propose first to split particles into groups with roughly the same radius. A separate set of models then needs to be trained for each group.   We  showed that training additional models requires much less effort since we can use transfer learning.

\section*{Acknowledgment}
\label{sec::acknowledgement}
The authors would like to thank C. Colonnello and L. Kondic for very useful comments on the manuscript. RS was supported by the Undergraduate Research Opportunity Program grant from the University of Oklahoma. MK was supported by a Junior Faculty Fellowship from the University of Oklahoma. 

\section*{References}
\bibliographystyle{iopart-num}
\bibliography{paper.bib}

\end{document}